\documentclass[conference]{IEEEtran}
\IEEEoverridecommandlockouts
\usepackage{cite}
\usepackage{amsmath,amssymb,amsfonts}
\usepackage{graphicx}
\usepackage{textcomp}
\usepackage{xcolor}
\usepackage{algorithm, algpseudocode}
\newcommand{\conds}{C}
\newcommand{\assign}{assign}

\def\BibTeX{{\rm B\kern-.05em{\sc i\kern-.025em b}\kern-.08em
    T\kern-.1667em\lower.7ex\hbox{E}\kern-.125emX}}
\begin{document}

\title{Metal Price Spike Prediction via a Neurosymbolic Ensemble Approach\\
}

\author{
\IEEEauthorblockN{Nathaniel Lee}
\IEEEauthorblockA{\textit{Arizona State University}\\
Tempe AZ 85281, USA \\
nlee51@asu.edu
}
\and
\IEEEauthorblockN{Noel Ngu}
\IEEEauthorblockA{\textit{Arizona State University} \\
Tempe AZ 85281, USA \\
nngu2@asu.edu
}
\and
\IEEEauthorblockN{Harshdeep Singh Sahdev}
\IEEEauthorblockA{\textit{Arizona State University} \\
Tempe AZ 85281, USA \\
hsahdev@asu.edu
}
\and
\IEEEauthorblockN{ Pramod Motaganahall}
\IEEEauthorblockA{\textit{Arizona State University} \\
Tempe AZ 85281, USA \\
pmotagan@asu.edu
}
\and
\IEEEauthorblockN{Al Mehdi Saadat Chowdhury}
\IEEEauthorblockA{\textit{Arizona State University} \\
Tempe AZ 85281, USA \\
achowd43@asu.edu
}
\and
\IEEEauthorblockN{Bowen Xi}
\IEEEauthorblockA{\textit{Arizona State University} \\
Tempe AZ 85281, USA \\
bowenxi@asu.edu
}
\and
\IEEEauthorblockN{Paulo Shakarian}
\IEEEauthorblockA{\textit{Arizona State University} \\
Tempe AZ 85281, USA \\
pshak02@asu.edu
}
}

\maketitle

\begin{abstract}
Predicting price spikes in critical metals such as Cobalt, Copper, Magnesium, and Nickel is crucial for mitigating economic risks associated with global trends like the energy transition and reshoring of manufacturing. While traditional models have focused on regression-based approaches, our work introduces a neurosymbolic ensemble framework that integrates multiple neural models with symbolic error detection and correction rules. This framework is designed to enhance predictive accuracy by correcting individual model errors and offering interpretability through rule-based explanations. We show that our method provides up to 6.42\% improvement in precision, 29.41\% increase in recall at 13.24\% increase in F1 over the best performing neural models.  Further, our method, as it is based on logical rules, has the benefit of affording an explanation as to which combination of neural models directly contribute to a given prediction.
\end{abstract}


\section{Introduction}
\label{sec:intro}
Global trends such as the energy transition from fossil fuels and the reshoring of manufacturing has led to an increased reliance on metals.  As a result, significant metal price fluctuation can lead to outsized economic impacts.  In this paper, we study the problem of predicting spikes in the prices of four critical metals: Cobalt, Copper, Magnesium, and Nickel.  While prior work \cite{ref_drachal_bayesian} \cite{ref_mohanty_agriculture} \cite{ref_kahraman_wavelet}  have mainly focused on regression and prediction by a single model, we focus on the identification of price spikes, which are potentially more economically destabilizing.  Further, we propose a neurosymbolic ensemble approach to enable the use of multiple models.  The way in which models are ensembeled is guided by an extension of the error detection and correction rules (EDCR), a symbolic rule learning technique~\cite{ref_bowen_edcr,kricheli2024errordetectionconstraintrecovery} (Section~\ref{sec:edcr}).  We find that our approach provides up to 6.42\% improvement in precision, 29.41\% increase in recall and 13.24\% increase in F1 over the best performing neural models (Table~\ref{table:model_eval}) in addition to an explanation as to which models contribute to a prediction that we explore with an ablation study (Section~\ref{sec:exp_ablation}).
\vspace{5pt}


\section{Related Work}
There have been several existing works on metal price prediction that rely on single model approaches, such as Support Vector Regression (SVR)~\cite{ref_svr_astudillo} and Artificial Neural Networks (ANNs)~\cite{ref_rapidminer_celik} to forecast general price trends by modeling non-linear relationships in metal price. While these models offer valuable insights, they may be limited in their ability to fully capture the complex patterns and variations in price movements. A multi-model approach can be more effective by combining the strengths of various models to more broadly capture price dynamics. 

Several studies have explored hybrid neural models that combine different architectures to enhance predictive power. For example, LSTM-CNN and LSTM-GRU models have been used to capture both short-term fluctuations and long-term trends in price data~\cite{ref_shi_forecast}. Another study combined LSTM and ANN with the GARCH model to predict copper price volatility, demonstrating the effectiveness of integrating statistical and deep learning methods~\cite{ref_hu_hybrid_lstmann}. Additionally, research has proposed hybrid models like GA-ELM and PSO-ELM, which optimize the parameters of Extreme Learning Machines (ELM) using meta-heuristic algorithms such as Genetic Algorithms (GA) and Particle Swarm Optimization (PSO), achieving more accurate copper price forecasts~\cite{ref_zheng_hybrid_elm}. These hybrid models have shown to outperform standalone models by leveraging complementary strengths across architectures.

While hybrid models integrate architectures within a single model, our ensemble approach offers further flexibility by combining multiple independent models, allowing us to capture a wider range of patterns and achieve more robust predictions. Additionally, recent studies have shown that self-attention mechanisms improve the ability to capture long-range dependencies in time-series data, leading to better performance in forecasting key trends over extended periods~\cite{ref_deepreg_zhou}. These mechanisms allow models to focus on important temporal features, making them particularly effective in time-series forecasting. Drawing from these insights, we have incorporated self-attention mechanisms into our neurosymbolic ensemble to enhance the detection of significant patterns and fluctuations, particularly in identifying metal price spikes.

\section{EDCR For Metal Price Prediction}
\label{sec:edcr}
In this paper, we define the Metal Price Spike Classification (MPSC) problem as the identification of significant shifts in price movements in the time series of metal commodity prices. Given a window of $n$ days, we wish to predict if the price at the ($n+1$)th day is ``anomalous'' which can be defined as either an unsupervised or semi-supervised approach.  For example, a spike can be a certain number of standard deviations above the moving average - hence this can be framed as a binary classification problem with two classes.  In our experiments, spikes are characterized as values that exceed or fall below the rolling mean by two rolling standard deviations, with a window size of 20 days. We note that other formulations to define spike are possible and that our approach does not depend on the specific definition of spikes.

Next, we describe the error detection and correction rule (EDCR) framework of \cite{ref_bowen_edcr} to address the MPSC problem.  Intuitively, given a trained model, \( f_{\theta} \), EDCR involves the learning of rules to detect and correct errors based on conditions under which the model is used.  These conditions can be aspects of the samples themselves or the environment. {A key intuition of this work is that \textit{we have one primary model and use other neural models as conditions} - as such EDCR provides customized rules that specify the best way to combine these models.  In this work, all the models are predicting the same binary class, as opposed to prior work where the primary model was always multi-class and the models for conditions were binary~\cite{ref_bowen_edcr} or where classes at different levels of granularity of the same model are used for conditions~\cite{kricheli2024errordetectionconstraintrecovery}.

\subsection{EDCR Rule Learning Algorithms (recap of \cite{ref_bowen_edcr})}
To effectively apply the EDCR framework for MPSC, we first need to understand the key rule-learning algorithms from \cite{ref_bowen_edcr}. These algorithms form the basis for detecting and correcting errors in the primary model's predictions.  The EDCR framework consists of two rule types: error detection and error correction. Each class in the dataset has one detection rule and one correction rule, designed to identify when the model's prediction is incorrect and how to correct it.\\

\noindent\textbf{Error Detection Rule Learning.}  The goal of error detection is to identify when a prediction made by the primary model \( f_\theta \) is incorrect. Algorithm~\ref{alg:negruleSelct}, \textsf{DetRuleLearn}~\cite{ref_bowen_edcr}, learns these rules by selecting conditions (denoted as \( DC_i \), the detection conditions for class \( i \)) that maximize precision while ensuring the recall reduction remains within a given threshold. The learned rule flags misclassifications by the primary model \( f_\theta \), which predicts a class \( pred_i(\omega) \), where \( \omega \) is a sample.

\begin{equation}
error(\omega) \leftarrow pred_i(\omega) \wedge \bigvee_{j \in DC_i} cond_j(\omega) \label{eq:error_detection}
\end{equation}

\begin{algorithm}[h]
\caption{\textsf{DetRuleLearn \cite{ref_bowen_edcr}}}
\label{alg:negruleSelct}
\begin{small}
\begin{algorithmic}[1]
    \Statex {\bfseries Require:} Class $i$, Recall reduction threshold $\epsilon$, Condition set $\conds$
    \Statex {\bfseries Ensure:} Subset of conditions $DC_i$
    \State{$DC_i:=\emptyset$}
    \State{$DC^* := \{ c \in \conds \textit{ s.t. } NEG_{\{c\}} \leq \epsilon \cdot \frac{N_iP_i}{R_i} \}$ }
    \While{$DC^* \neq \emptyset$}
        \State{$c_{best}=\arg\max_{c \in DC^*} POS_{DC_i\cup\{c\}}$}
        \State{Add $c_{best}$ to $DC_i$}
        \State{$DC^* :=\{ c \in \conds\setminus DC_i\textit{ s.t. }  NEG_{DC_i\cup\{c\}} \leq \epsilon \cdot \frac{N_iP_i}{R_i} \}$}
    \EndWhile
    \State{\textbf{return} $DC_i$}
\end{algorithmic}
\end{small}
\end{algorithm}

Here, \( POS_{DC} \) is the number of samples where the conditions are met and the model's prediction is an error, indicating how well the detection rule identifies misclassifications. \( NEG_{DC} \) counts the samples where the conditions are met, but the prediction was correct. \( BOD \) represents all samples that satisfy the rule's conditions, regardless of whether the prediction was correct. The \textsf{DetRuleLearn} algorithm iteratively selects the best condition $c_{best}$ by maximizing \( POS_{DC} \) while ensuring \( NEG_{DC} \) does not exceed the recall reduction threshold $\epsilon$. Each iteration evaluates each condition in $\conds$, which leads to a time complexity of $O(|\conds|^2 \cdot N)$, where $N$ is the number of training samples. 
 The use of data structures can likely reduce this further.\\

\noindent\textbf{Error Correction Rule Learning.}  Once a misclassification is detected, Algorithm~\ref{alg:posRuleSelct}, \textsf{CorrRuleLearn}~\cite{ref_bowen_edcr} algorithm reassigns the sample to the correct class. It does this by selecting condition-class pairs (denoted as \( CC_i \), the prior assigned class and associated condition that allows us to reclassify the sample to  \( i \)) to maximize both precision and recall. The correction rules are based on the output of other models or conditions, represented as \( cond_q(\omega) \), which help correct the primary model's prediction.

\begin{equation}
corr_i(\omega) \leftarrow \bigvee_{q,r \in CC_i} (cond_q(\omega) \wedge pred_r(\omega)) \label{eq:correction_rule}
\end{equation}

\begin{algorithm}[h]
\caption{\textsf{CorrRuleLearn \cite{ref_bowen_edcr}}
\label{alg:posRuleSelct}}
\begin{small}
\begin{algorithmic}[1]
\Statex {\bfseries Require:} Class $i$, Set of condition-class pairs $CC_{all}$
\Statex {\bfseries Ensure:} Subset of condition-class pairs $CC_i$
\State{$CC_i := \emptyset$}
\State{$CC_i' := CC_{all}$}
\State{Sort each $(c,j)\in CC_{all}$ from greatest to least by $\frac{POS_{\{(c,j)\}}}{BOD_{\{(c,j)\}}}$ and remove $\frac{POS_{\{(c,j)\}}}{BOD_{\{(c,j)\}}} \leq P_i $ }

\For{$(c,j) \in CC_{all}$ selected in order of the sorted list}
    \State $a := \frac{POS_{CC_i \cup \{(c,j)\}}}{BOD_{CC_i\cup \{(c,j)\}}}-\frac{POS_{CC_i}}{BOD_{CC_i}}$
    \State $b := \frac{POS_{CC_i' \setminus \{(c,j)\}}}{BOD_{CC_i' \setminus \{(c,j)\}}}-\frac{POS_{CC_i'}}{BOD_{CC_i'}}$
    \If{$a \geq b$}
        \State{$CC_i := CC_i \cup \{(c,j)\}$}
    \Else
        \State{$CC_i':= CC_i' \setminus \{(c,j)\}$}
    \EndIf
\EndFor

\If{$\frac{POS_{CC_i}}{BOD_{CC_i}}\leq P_i$}
    \State{$CC_i := \emptyset$}
\EndIf
\State \textbf{return} $CC_i$
\end{algorithmic}
\end{small}
\end{algorithm}

The algorithm begins by sorting the condition-class pairs in $CC_{all}$ based on their precision ratios. After sorting, it iteratively evaluates each pair to determine whether adding it to $CC_i$ improves the correction rule. This iterative evaluation continues until the best subset of condition-class pairs is identified. The time complexity of \textsf{CorrRuleLearn} is \(O(N\cdot m+ m \log m)\), where $m$ is the number of condition-class pairs in $CC_{all}$ and $N$ is the number of samples. This complexity is primarily due to the sorting step, followed by a linear pass through the pairs (that involves the $O(N)$ operations of recalculating $POS$ and $BOD$) to finalize the correction set.

\subsection{EDCR Applied to MPSC}  

As with \cite{ref_bowen_edcr,kricheli2024errordetectionconstraintrecovery} model \( f_{\theta} \), the conditions, and the EDCR rules are created based on the training data.  We instantiate the logical language for our use case as follows: $\assign_{spike},\assign_{no}$ meaning that $f_\theta$ assigned a given sample class $spike$ (if a price spike is predicted) or $no$ (for no spike predicted), $corr_{spike}$ meaning that the model's prediction should be corrected to class $spike$ (note due to the binary nature of the problem, we do not use the corresponding $corr_{no}$ predicate in this work), and $cond$ meaning that condition $cond$ is true for the given sample.  In this work, for example, $cond_{CNN1}$ would mean that a model ``CNN1'' (i.e. CNN variant 1 as per Table ~\ref{table:model_configurations}) -- which is not $f_\theta$ -- assigned a sample as having a price spike.  We show two example rules obtained from our modified EDCR rule learner in Table~\ref{tab:example_rls_Rules}.  Notice that here our conditions are based directly on the classification decisions of other models.

\begin{table}[]
\scriptsize
\centering
\caption{Example EDCR Rule Learned for the MPSC Problem}
\label{tab:example_rls_Rules}
\begin{tabular}{@{}p{0.1\textwidth}p{0.3\textwidth}}
\hline
Rule & Machine Learning Models\\
\hline
$corr_{spike}(w) \leftarrow \hfill \assign_{no}(w) \wedge cond_{CNN1}(w) $ &
If base model \( f_{\theta} \) identifies sample $w$ as not having a spike (class $no$) and auxiliary model CNN1 classifies a certain day in the time series as a $spike$, then the overall ensemble prediction is $spike$. \\\\

\hline 

$corr_{spike}(w) \leftarrow \assign_{no}(w) \wedge cond_{RNN4}(w) $ &
If base model \( f_{\theta} \) identifies sample $w$ as not having a spike (class $no$) and auxiliary model RNN4 classifies a certain day in the time series as a $spike$, then the overall ensemble prediction is $spike$.  \\
\\

\hline
\end{tabular}
\end{table}

\subsection{Modified EDCR Rule Learner for MPSC}  
In this paper we extend the work of \cite{ref_bowen_edcr} with an overall approach suited for the MPSC problem called \textsf{MPSCRuleLearn} that calls both \textsf{DetRuleLearn} and \textsf{CorrRuleLearn} as subroutines.  In this adaptation, both the detection and correction rules lead directly to corrective actions due to the binary nature of the classification problem. This eliminates the need to separately define error detection and correction rules, simplifying the learning process.  Furthermore, we introduced a condition selection method called Top F1, where conditions are chosen based on their performance during training. By prioritizing conditions that yield the highest F1 scores, we ensure that only the most effective conditions are included in the rule learning process. This helps maintain high rule quality by filtering out underperforming conditions that could otherwise negatively impact the results. We note that \textsf{MPSCRuleLearn}  simply makes calls to \textsf{DetRuleLearn} and \textsf{CorrRuleLearn}, so the runtime is additive of the two.

\begin{algorithm}
\caption{\textsf{MPSCRuleLearn}\label{alg:ruleSelct}}
\begin{algorithmic}[1]
\Require Recall reduction threshold $\epsilon$, Condition set $\conds$
\Ensure Subset of conditions $DC_i$

\State {$\conds_f$ := Filtered Condition set C.}
\State $CC_{all} := \{\forall_c \in \conds_f : (nospike, c)\} $
\State $DC = \textsf{DetRuleLearn}(nospike, \epsilon, \conds_f) $
\State $CC = \textsf{CorrRuleLearn}(spike, CC_{all}) $

\State \textbf{return} \{
\begin{align*}
    & \quad corr_{\text{spike}}(w) \leftarrow pred_{\text{no}}(w) \wedge \bigvee_{cond \in DC} cond(w), \\
    & \quad corr_{\text{spike}}(w) \leftarrow \bigvee_{cond \in CC}(pred_{\text{no}}(w) \wedge cond(w))
\end{align*}
\}

\end{algorithmic}
\end{algorithm}

\section{Experiments}
\label{sec:experiments}

In this section, we describe a set of experiments where we evaluated our approach on price spike prediction for four metals (Cobalt, Copper, Magnesium, and Nickel).  We also describe our experimental setup as well as an ablation study.

\subsection{Experimental Settings} 

\paragraph{Dataset and Setup} In our experiments, we utilized datasets of daily open prices for four distinct metals—cobalt, Copper, Magnesium, and Nickel—from \cite{ref_investing_com}, spanning the years 2020 to 2023. This range of metals was selected to assess the applicability of our results across diverse commodity markets. Each row in the dataset contains the open, high, and low prices of a metal on a specific day. The datasets contain 1039 samples of cobalt, 1008 samples of Copper, 716 samples of Magnesium, and 1006 samples of Nickel. Datasets were split into training and test sets at a 60:40 ratio across all datasets. The datasets were first sorted by their date. Then, the first 60\% of the dataset was placed into the training set while the last 40\% was put into the test set. \\
\paragraph{Machine Learning Models} In our experiments, we trained and fine-tuned a range of machine-learning models. Specifically, we employed Convolutional Neural Networks (CNN), CNN with Attention mechanism (ATTN), Long Short-Term Memory (LSTM) networks, and Recurrent Neural Networks (RNN).  In total, there were 4 variants of RNNs and LSTMs, and 12 variants of CNN and ATTN. These variants are elaborated on in Table ~\ref{table:model_configurations}.\\
\paragraph{Hardware and Implementation} Experiments were performed on a AMD Ryzen 9 7420HS CPU, Radeon 780M GPU using Python 3.11 with Tensorflow. \\

\subsection{Model Evaluation}
\label{sec:exp_models}

We evaluated how effectively the EDCR framework improved the performance of our foundational classification models: CNN, RNN, LSTM, and ATTN. Each model was trained using features extracted from the daily price data of various metals. After generating a pool of classification results with various model configurations and hyperparameters, these results were processed using the EDCR framework, using our modified Algorithm \ref{alg:ruleSelct}. 
Table \ref{table:model_eval} presents the best-performing models for each metal dataset, based on precision, recall, and F1 scores, along with the improvements made by applying EDCR. For the condition selection method (Top F1), we selected the top 200 conditions (there were 744 in total) based on F1 score to apply the EDCR corrections. Random Baseline Performance when Recall is set to 1 (e.g., predict spikes for all input) for metals Cobalt, Copper, Magnesium, and Nickel are 0.15, 0.14, 0.19, 0.14 respectively. The base models shown in the table for each metal represent the best precision, best recall, best F1 among all models trained.  In general, a baseline model augmented with EDCR improved in all metrics over the baseline model - with the exception of high-precision baseline models.  Compared to all trained models, recall and F1 were both significantly improved with the use of EDCR (of note F1 improved by over $10\%$ in three of four metals).  We note that our approach is designed to increase the recall of the $spike$ class (while minimizing recall decrease to the non-spike class) - which is recall improvements seem to drive the results in most cases (and also why in only one of four metals precision improved significantly over the baseline).  However, it is noteworthy that in all but one case, the top recall or top F1 model also improved or maintained precision over its non-EDCR counterpart.

\newcommand{\metalseparator}[1]{\hline \multicolumn{4}{|c|}{#1}\\ \hline}
\newcommand{\tablelabels}{Model Variant & Precision & Recall & F1 \\ \hline }
\newcommand{\emptyrow}{\multicolumn{4}{c}{} \\}

\begin{table*}[h!]
\caption{Model Evaluation Results for Cobalt, Copper, Magnesium and Nickel. Using the MPSCRuleLearn algorithm with Top F1 filtering. The base models (no EDCR) with the best precision, recall and F1 are each underlined. The overall best performing models in terms of precision, recall, F1 across all models are bolded.}
\centering
\label{table:model_eval}

\begin{minipage}{0.48\textwidth}
\centering
\begin{tabular}{|c|c|c|c|}
\metalseparator{Cobalt}\tablelabels
CNN  (12) & 0.82 & \underline{0.65} & \underline{0.73}\\
ATTN (8) & \underline{\textbf{0.94}} & 0.19 & 0.32\\
\hline
CNN  (12) (EDCR) & 0.80 (-2.15\%) & \textbf{0.85} (+29.41\%) & \textbf{0.83 (+13.24\%)}\\
ATTN (8) (EDCR) & \textbf{0.94} (0.0\%) & 0.19 (0.0\%) & 0.32 (0.0\%)\\
\hline
\emptyrow
\emptyrow
\emptyrow
\metalseparator{Copper}\tablelabels
ATTN (2) & 0.78 & 0.74 & \underline{0.76}\\
CNN  (3) & 0.52 & \underline{0.78} & 0.62\\
CNN  (4) & \underline{\textbf{0.83}} & 0.52 & 0.64\\
\hline
ATTN (2) (EDCR) & 0.79 (+0.5\%) & 0.76 (+2.33\%) & \textbf{0.77} (+1.43\%)\\
CNN  (3) (EDCR) & 0.53 (+0.69\%) & \textbf{0.84} (+8.89\%) & 0.65 (+3.84\%)\\
CNN  (4) (EDCR) & \textbf{0.83} (0.0\%) & 0.52 (0.0\%) & 0.64 (0.0\%)\\
\hline
\emptyrow
\end{tabular}
\end{minipage}
\hfill
\begin{minipage}{0.48\textwidth}
\centering
\begin{tabular}{|c|c|c|c|}
\metalseparator{Magnesium}\tablelabels
CNN  (1) & 0.52 & 0.65 & \underline{0.58}\\
RNN  (4) & 0.20 & \underline{0.99} & 0.33\\
CNN  (2) & \underline{\textbf{0.86}} & 0.17 & 0.28\\
\hline
CNN  (1) (EDCR) & 0.53 (+3.19\%) & 0.79 (+21.74\%) & \textbf{0.64} (+10.67\%)\\
RNN  (4) (EDCR) & 0.20 (+0.87\%) & \textbf{1.00} (+1.43\%) & 0.33 (+0.96\%)\\
CNN  (2) (EDCR) & \textbf{0.86} (0.0\%) & 0.17 (+0.0\%) & 0.28 (0.0\%)\\
\hline
\emptyrow
\metalseparator{Nickel}\tablelabels
ATTN (5) & 0.59 & 0.48 & \underline{0.53}\\
CNN  (6) & 0.37 & \underline{0.57} & 0.45\\
ATTN (8) & \underline{0.66} & 0.38 & 0.48\\
\hline
ATTN (5) (EDCR) & 0.60 (+1.38\%) & 0.59 (+24.14\%) & \textbf{0.60} (+12.85\%)\\
CNN  (6) (EDCR) & 0.40 (+6.42\%) & \textbf{0.69} (+20.0\%) & 0.50 (+11.38\%)\\
ATTN (8) (EDCR) & \textbf{0.68} (+2.82\%) & 0.41 (+8.7\%) & 0.51 (+6.48\%)\\
\hline
\emptyrow
\end{tabular}
\end{minipage}
\end{table*}

\subsection{Ablation Studies}
\label{sec:exp_ablation}

We studied the effects of ablating different models from the condition set  $\conds$ used by the EDCR algorithm, where one model's output corrects another under different hyperparameters or architectures, impacting the precision and recall of the base models. Our findings indicate that excluding LSTM and RNN-based results from the condition set  $\conds$ generally has little to no effect on overall model performance, supporting our initial assumption that they would not contribute to improved overall performance of the ensemble. Specifically, our ablation analysis in Figure \ref{fig:ablation1} on the CNN (1) model (left chart) for predicting Magnesium price spikes showed that removing CNN Attention from the conditions resulted in a drop in both Precision (-1\%) and Recall (-13\%) while removing other configurations of CNN only caused a drop in Recall (6\%). Similarly, for the CNN (5) model (right chart) used in predicting Cobalt price spikes, the exclusion of both differently configured CNN and CNN Attention from the conditions led to moderate decreases in Recall (-19\% and -18\% respectively), with slight increases in Precision (+2\%). It should be noted that in both these situations where a CNN was used as the base model and other CNN configurations were excluded, there was a performance decline, indicating that these variations could detect price spikes that the base model missed.
\begin{figure}[hbt!]
    \centering
    
    \includegraphics[width=0.5\textwidth]{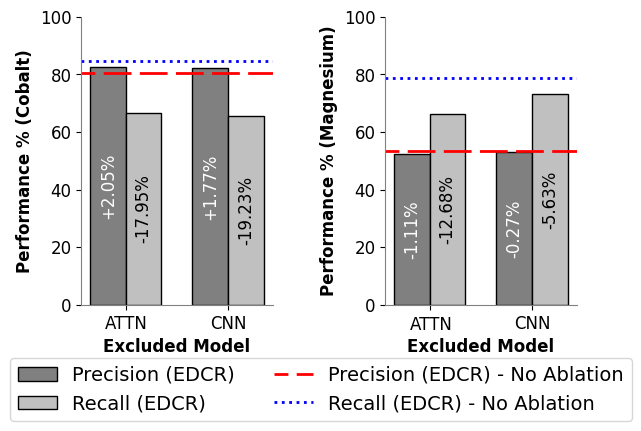}
    \caption{Ablation analysis of EDCR on ATTN (1)  and model (left) and CNN (5) model (right) performance metrics for predicting Magnesium and Cobalt price spikes respectively. The bar charts demonstrate varying Precision and Recall tradeoffs when different models are used as rules for the base model.}
    
    \label{fig:ablation1}
\end{figure}



\section{Conclusion}

In this paper, we use a variant of error detection and correction rules (EDCR) to ensemble multiple neural models to predict spikes in the prices of various metals.  Using this technique, we obtained significant improvements over a single model as we found through ablation studies that there is variation on the performance of individual models - particularly impacting recall. While the current study focuses on metals like Cobalt, Magnesium, Copper, and Nickel, future work will explore extending the method to other metals, such as Lithium, and potentially to broader commodities like agricultural or energy products, to enhance generalizability. Another direction for future work is moving beyond the current focus on predicting spikes in metal prices (classification) and instead using regression to forecast the actual price. Shifting to a regression framework would allow us to conduct a more detailed comparative analysis with recent state-of-the-art approaches ~\cite{ref_svr_astudillo, ref_rapidminer_celik, ref_shi_forecast, ref_hu_hybrid_lstmann, ref_zheng_hybrid_elm}. Additionally, a shortcoming of this paper is that the results from applying EDCR results are deterministic. As such, probabilistic semantics is another direction for future work.

\section{Acknowledgment}

Some of the authors are supported by ONR grant \\
N00014-23-1-2580 and ARO grant W911NF-24-1-0007.

\section{Appendix}
\newcommand{\modelseparator}[1]{\hline \multicolumn{2}{|c|}{#1}\\ \hline} 
\newcommand{\tablelabelsRNN}{Layers \\ \hline}  
\newcommand{\tablelabelsCNN}{Filter Size & Kernel Size \\ \hline}  
\newcommand{\emptyrowmodel}{\multicolumn{2}{c}{} \\} 

Table \ref {table:model_configurations} contains the various hyperparameters that were applied across all variations of each classification model.

\begin{table}[hbt!]
\caption{Variations of specific hyperparameters across the classification models used in the experiments.}
\centering
\label{table:model_configurations}

\begin{minipage}{0.48\columnwidth}
\centering
\begin{tabular}{|c|c|}
\modelseparator{RNN and LSTM}
{Variant} & {Layers} \\
\hline
1 & 32 \\
2 & 64 \\
3 & 128 \\
4 & 256 \\
\hline
\emptyrowmodel
\end{tabular}
\end{minipage}
\hfill
\begin{minipage}{0.48\columnwidth}
\centering
\begin{tabular}{|c|c|}
\modelseparator{CNN and ATTN}
{Variant} & {Filters, Kernel Size} \\
\hline
1 & 32, 7 \\
2 & 32, 5 \\
3 & 32, 3 \\
4 & 64, 7 \\
5 & 64, 5 \\
6 & 64, 3 \\
7 & 128, 7 \\
8 & 128, 5 \\
9 & 128, 3 \\
10 & 256, 7 \\
11 & 256, 5 \\
12 & 256, 3 \\
\hline
\end{tabular}
\end{minipage}

\end{table}

\pagebreak
\bibliographystyle{IEEEtran}
\bibliography{metals}

\end{document}